\documentstyle[conll98]{article}

\input{epsf}    
\input{ipamacs} 

\title{Do Not Forget: \\ Full Memory in Memory-Based Learning of Word
Pronunciation 
\thanks{This research was done in the context of the ``Induction of
Linguistic Knowledge'' research programme, partially supported by the
Foundation for Language Speech and Logic (TSL), which is funded by the
Netherlands Organization for Scientific Research (NWO). Part of the
first author's work was performed at the Department of Computer
Science of the Universiteit Maastricht.}}
\author{Antal van den Bosch \and Walter Daelemans \\
Tilburg University, ILK \\ P.O. Box 90153,
NL-5000 LE Tilburg \\ The Netherlands \\ {\tt
\{antalb,walter\}@kub.nl}}
\begin{document}
\maketitle

\begin{abstract}
Memory-based learning, keeping full memory of learning material,
appears a viable approach to learning {\sc nlp} tasks, and is often
superior in generalisation accuracy to eager learning approaches that
abstract from learning material. Here we investigate three {\em
partial}\/ memory-based learning approaches which remove from memory
specific task instance types estimated to be exceptional. The three
approaches each implement one heuristic function for estimating
exceptionality of instance types: (i) typicality, (ii) class
prediction strength, and (iii) friendly-neighbourhood
size. Experiments are performed with the memory-based learning algorithm
{\sc ib1-ig} trained on English word pronunciation. We find that
removing instance types with low prediction strength (ii) is the only
tested method which does not seriously harm generalisation
accuracy. We conclude that keeping full memory of types rather
than tokens, and excluding minority ambiguities appear to be the only
performance-preserving optimisations of memory-based learning.
\end{abstract}

\section{Introduction}

Memory-based learning of classification tasks is a branch of
supervised machine learning in which the learning phase consists
simply of storing all encountered instances from a training set in
memory \cite{aha-lazybook}. Memory-based learning algorithms do not
invest effort during learning in abstracting from the training data,
such as eager-learning (e.g., decision-tree algorithms,
rule-induction, or connectionist-learning algorithms,
\cite{quinlan-c45,mitchell-book}) do. Rather, they defer investing
effort until new instances are presented. On being presented with an
instance, a memory-based learning algorithm searches for a
best-matching instance, or, more generically, a set of the $k$
best-matching instances in memory. Having found such a set of $k$
best-matching instances, the algorithm takes the (majority) class with
which the instances in the set are labeled to be the class of the new
instance.  Pure memory-based learning algorithms implement the classic
$k$-nearest neighbour algorithm \cite{cover-knn,devijver-knn,aha-ibl}; in
different contexts, memory-based learning algorithms have also been
named lazy, instance-based, exemplar-based, memory-based, case-based
learning or reasoning
\cite{stanfill-nettalk,kolodner-cbr,aha-ibl,aha-lazybook})

Memory-based learning has been demonstrated to yield accurate models of
various natural language tasks such as grapheme-phoneme conversion,
word stress assignment, part-of-speech tagging, and PP-attachment
\cite{daelemans-overview}. For example, the memory-based learning algorithm
{\sc ib1-ig} \cite{daelemans-hyphen,daelemans-air}, which extends the
well-known {\sc ib1} algorithm \cite{aha-ibl} with an information-gain
weighted similarity metric, has been demonstrated to perform
adequately and, moreover, consistently and significantly better than
{\em eager-learning}\/ algorithms which do invest effort in
abstraction during learning (e.g., decision-tree
learning \cite{daelemans-air,quinlan-c45}, and connectionist learning
\cite{rumelhart-bp}) when trained and tested on a range of
morpho-phonological tasks (e.g., morphological segmentation,
grapheme-phoneme conversion, syllabification, and word stress
assignment) \cite{daelemans-cl,vandenbosch-morph,vandenbosch-thesis}.
Thus, when learning {\sc nlp} tasks, the
abstraction occurring in decision trees (i.e., the explicit {\em
forgetting}\/ of information considered to be redundant) and in
connectionist networks (i.e., a non-symbolic encoding and decoding
in relatively small numbers of connection weights) both hamper
accurate generalisation of the learned knowledge to new material.

These findings appear to contrast with the general assumption behind eager
learning, that data representing real-world classification tasks tends
to contains (i) redundancy and (ii) exceptions: redundant data can be removed,
yielding smaller descriptions of the original data; some exceptions
(e.g., low-frequency exceptions) can (or should) be discarded since
they are expected to be bad predictors for classifying new (test)
material. However, both redundancy and exceptionality cannot be
computed trivially; heuristic functions are generally used to estimate
them (e.g., functions from information theory \cite{quinlan-c45}). The
lower generalisation accuracies of both decision-tree and
connectionist learning, compared to memory-based learning, on the
above-mentioned {\sc nlp} tasks, suggest that these heuristic
estimates may not hold for data representing {\sc nlp} tasks. It
appears that in order to learn such tasks successfully, a learning
algorithm should not forget (i.e., explicitly remove from memory) any
information contained in the learning material: it should not abstract
from the individual instances.

An obvious type of abstraction that is not harmful for generalisation
accuracy (but that is not always acknowledged in implementations of
memory-based learning) is be the straightforward abstraction
from tokens to types with frequency information.  In general, data
sets representing natural language tasks, when large enough, tend to
contain considerable numbers of duplicate sequences mapping to the
same output or class. For example, in data representing word
pronunciations, some sequences of letters, such as {\sf ing} at the
end of English words, occur hundreds of times, while each of the
sequences is pronounced identically, viz. /\i \eng /. Instead of
storing all individual sequence tokens in memory, each set of
identical tokens can be safely stored in memory as a single sequence
type with frequency information, without loss of generalisation
accuracy \cite{daelemans-hyphen,daelemans-air}. Thus, forgetting
instance tokens and replacing them by instance types may lead to
considerable computational optimisations of memory-based learning,
since the memory that needs to be searched may become considerably
smaller.

Given the safe, performance-preserving optimisation of replacing sets
of instance tokens by instance types with frequency information, a
next step of investigation into optimising memory-based learning is to measure
the effects of {\em forgetting instance types}\/ on grounds of their
exceptionality, the underlying idea being that the more exceptional a
task instance type is, the more likely it is that it is a bad predictor for
new instances. Thus, exceptionality should in some way express the
unsuitability of a task instance type to be a best match (nearest
neighbour) to new instances: it would be unwise to copy its associated
classification to best-matching new instances. In this paper, we
investigate three criteria for estimating an instance type's
exceptionality, and removing instance types estimated to be
the most exceptional by each of these criteria. The criteria investigated are

\begin{enumerate}
\item typicality of instance types;
\item class prediction strength of instance types;
\item friendly-neighbourhood size of instance types;
\item random (to provide a baseline experiment).
\end{enumerate}

We base our experiments on a large data set of English word
pronunciation. We briefly describe this data set, and the way it is
converted into an instance base fit for memory-based learning, in
Section~\ref{data}. In Section~\ref{experimental} we describe
the settings of our experiments and the memory-based learning algorithm {\sc
ib1-ig} with which the experiments are performed. We then turn to
describing the notions of typicality, class-prediction strength, and
friendly-neighbourhood size, and the functions to estimate them, in
Section~\ref{criteria}.  Section~\ref{results} provides the
experimental results. In Section~\ref{discussion}, we discuss the
obtained results and formulate our conclusions.

\section{The word-pronunciation data}
\label{data}

Converting written words to stressed phonemic transcription, i.e.,
word pronunciation, is a well-known benchmark task in machine learning
\cite{stanfill-nettalk,sejnowski-nettalk,dietterich-id3,wolpert-nettalk}.  We
define the task as the conversion of fixed-sized instances
representing parts of words to a class representing the phoneme and
the stress marker of the instance's middle letter. To generate the
instances, windowing is used \cite{sejnowski-nettalk}.
Table~\ref{overall-windows-ex} displays example instances and their
classifications generated on the basis of the sample word {\sf
booking}. Classifications, i.e.,  phonemes with stress markers
(henceforth PSs), are denoted by
composite labels. For example, the first instance in
Table~\ref{overall-windows-ex}, {\sf \_\_\_book}, maps to class label
/b/1, denoting a /b/ which is the first phoneme of a syllable
receiving primary stress. In this study, we chose a fixed window width
of seven letters, which offers sufficient context information for
adequate performance, though extension
of the window decreases ambiguity within the data set
\cite{vandenbosch-thesis}. 

\begin{table*}
\begin{center}
\begin{tabular}{||c||ccc|c|ccc||c||}
\hline
instance & \multicolumn{3}{|c|}{left} & focus &
\multicolumn{3}{|c||}{right} &
classification \\
number & \multicolumn{3}{|c|}{context} & letter &
\multicolumn{3}{|c||}{context} & \\
\hline
1  & \_ & \_ & \_ & {\sf b} & {\sf o} & {\sf o} & {\sf k}      & /b/1 \\
2  & \_ & \_ & {\sf b} & {\sf o} & {\sf o} & {\sf k} & {\sf i} & /u/0 \\
3  & \_ & {\sf b} & {\sf o} & {\sf o} & {\sf k} & {\sf i} & {\sf n} & /-/0 \\
4  & {\sf b} & {\sf o} & {\sf o} & {\sf k} & {\sf i} & {\sf n} & {\sf g} & /k/0 \\
5  & {\sf o} & {\sf o} & {\sf k} & {\sf i} & {\sf n} & {\sf g} & \_ &
/\i /0 \\
6  & {\sf o} & {\sf k} & {\sf i} & {\sf n} & {\sf g} & \_ & \_ & /\eng /0 \\
7  & {\sf k} & {\sf i} & {\sf n} & {\sf g} & \_ & \_ & \_ & /-/0 \\
\hline
\end{tabular}
\normalsize
\caption{Example of instances generated for the word-pronunciation
task from the word {\sf booking}.\label{overall-windows-ex}}
\end{center}
\end{table*}

The task, henceforth
referred to as {\sc gs} ({\sc g}rapheme-phoneme conversion and {\sc
s}tress assignment) is similar to
the {\sc nettalk} task presented by Sejnowski and Rosenberg (1986),
but is performed on a larger corpus of 77,565 English
word-pronunciation pairs, extracted from the {\sc celex} lexical data
base \cite{burnage-celex}. Converted into
fixed-sized instance, the full instance base representing the {\sc gs}
task contains 675,745 instances. The task features 159 classes
(combined phonemes and stress markers).

\section{Algorithm and experimental setup}
\label{experimental}

\subsection{Memory-based learning in IB1-IG}

In the experiments reported here, we employ {\sc ib1-ig}
\cite{daelemans-hyphen,daelemans-air}, which has been demonstrated to
perform adequately, and better than eager-learning algorithms on the
{\sc gs} task \cite{vandenbosch-thesis}. {\sc
ib1-ig} constructs an instance base during learning. An instance in
the instance base consists of a fixed-length vector of $n$
feature-value pairs (here, $n=7$), an information field containing the
classification of that particular feature-value vector, and an
information field containing the occurrences of the instance with its
classification in the full training set. The latter information field
thus enables the storage of instance types rather than the more
extensive storage of identical instance tokens. After the instance
base is built, new (test) instances are classified by matching them to
all instance types in the instance base, and by calculating with each match
the {\em distance}\/ between the new instance $X$ and the memory
instance type $Y$, $\Delta(X,Y)$, using the function given in
Eq.~\ref{simfunc}:
\begin{equation}
\Delta(X,Y)=\sum_{i=1}^{n}W(f{_i})\delta(X_{i},Y_{i}),
\label{simfunc}
\end{equation}
where $W(f{i})$ is the weight of the
$i$th feature,
and $\delta(x_{i},y_{i})$ is the distance 
between the values of the $i$th feature in the instances $X$ and $Y$.
When the values of the instance features are symbolic, as with the
{\sc gs} task (i.e., feature values are letters), a simple distance
function is used (Eq. \ref{simpdist}): 
\begin{equation}
\delta(X_{i},Y_{i}) = 0 \; if \; X_{i}=Y_{i} \; else \; 1.
\label{simpdist}
\end{equation}
The classification of the memory instance type 
$Y$ with the smallest $\Delta(X,Y)$ is then taken as the classification
of $X$. This procedure is also known as {\sc 1-nn}, i.e., a search for
the single nearest neighbour, the simplest variant of $k$-{\sc nn} 
\cite{devijver-knn}. 

The weighting function of {\sc ib1-ig}, $W(f{_i})$, represents the
{\em information gain}\/ of feature $f_{i}$. Weighting features in
$k$-{\sc nn} classifiers such as {\sc ib1-ig} is an active field of
research (cf. \cite{wettschereck-thesis,wettschereck-weighting}, for
comprehensive overviews and discussion). Information gain is a
function from information theory also used in {\sc id3}
\cite{quinlan-id3} and {\sc c4.5} \cite{quinlan-c45}.  The information
gain of a feature expresses its relative relevance compared to the
other features when performing the mapping from input to
classification.

The idea behind computing the information gain of features is to
interpret the training set as an information source capable of
generating a number of messages (i.e., classifications) with a certain 
probability. The information entropy $H$ of such an information source can
be compared in turn for each of the features characterising the
instances (let $n$ equal the number of features), to the average information
entropy of the information source when the value of those features are
known.
 
Data-base information entropy $H(D)$ is equal to the number of bits of
information needed to know the classification given an instance. It is
computed by equation~\ref{dbentropy}, where $p_{i}$ (the probability
of classification $i$) is estimated by its relative frequency in the
training set. 
 
\begin{equation}
H(D) = - \sum_{i} p_{i} log_{2} p_{i}
\label{dbentropy}
\end{equation}
 
To determine the information gain of each of the $n$ features $f_{1}
\ldots f_{n}$, we compute the average information entropy for each
feature and subtract it from the information entropy of the data
base. To compute the average information entropy for a feature
$f_{i}$, given in equation~\ref{featentropy}, we take the average
information entropy of the data base restricted to each possible value
for the feature. The expression $D_{[f_{i}=v_{j}]}$ refers to those
patterns in the data base that have value $v_{j}$ for feature $f_{i}$,
$j$ is the number of possible values of $f_{i}$, and $V$ is the set of
possible values for feature $f_{i}$. Finally, $|D|$ is the number of
patterns in the (sub) data base.

\begin{equation}
H(D_{[f_{i}]}) = \sum_{v_{j} \in V} H(D_{[f_{i}=v_{j}]}) \frac{|D_{[f_{i}=v_{j}]}|}{|D|}
\label{featentropy}
\end{equation}

Information gain of feature $f_{i}$ is then obtained by
equation~\ref{funcinfogain}. 

\begin{equation}
G(f_{i}) = H(D) - H(D_{[f_{i}]})
\label{funcinfogain}
\end{equation}

Using the weighting function $W(f{_i})$ acknowledges the fact that for
some tasks, such as the current {\sc gs} task, some features are far
more relevant (important) than other features. Using it, instances
that match on a feature with a relatively high information gain are
regarded as less distant (more alike) than instances that match on a
feature with a lower information gain.

Finding a nearest neighbour to a test instance may result in two or
more candidate nearest-neighbour instance types at an identical
distance to the test instance, yet associated with different
classes. The implementation of {\sc ib1-ig} used here handles such
cases in the following way. First, when two or more best-matching instance
types are found associated to different classes, {\sc ib1-ig} selects
the class of the instance type with the highest occurrence in the
set of best-matching instance types. In case of occurrence ties, the
classification of one of the set of best-matching instance types is
selected that has the highest overall
occurrence in the training set.
\cite{daelemans-air}.

\subsection{Setup}

We performed a series of experiments in which {\sc ib1-ig} is applied
to the {\sc gs} data set, systematically edited according to each of
the three tested criteria (plus the baseline random criterion)
described in the next section. We performed the following global 
procedure:

\begin{enumerate}
\item
We partioned the full {\sc gs} data set into a training set of 608,228
instances (90\% of the full data set) and a test set of 67,517
instances (10\%). For use with {\sc ib1-ig}, which stores instance
types rather than instance tokens, the data set was reduced to contain
222,601 instance types (i.e., unique combinations of feature-value vectors and
their classifications), with frequency information.
\item
For each exceptionality criterion (i.e.,
typicality, class prediction strength, friendly-neighbourhood size,
and random selection),
	\begin{enumerate}
	\item we created four edited instance bases by removing 1\%, 2\%,
	5\%, and 10\% of the most exceptional instance types (according to
	the criterion) from the training set, respectively.
	\item For each of these increasingly edited training sets,
	we performed one experiment in which {\sc ib1-ig} was trained
	on the edited training set, and tested on the original
	unedited test set.
	\end{enumerate}
\end{enumerate}

\section{Three estimations of exceptionality}
\label{criteria}

We investigate three methods for estimating the (degree of)
exceptionality of instance types: typicality, class prediction
strength, and friendly-neighbourhood size. 

\subsection{Typicality}

In its common meaning, ``typicality'' denotes roughly the opposite
of exceptionality; atypicality can be said to be a synonym of
exceptionality.  We adopt a definition from \cite{zhang-typical}, who
proposes a function to this end. Zhang computes typicalities of
instance types by taking both their feature values and their
classifications into account \cite{zhang-typical}. He adopts the
notions of {\em intra-concept similarity}\/ and {\em inter-concept
similarity}\/ \cite{rosch-family} to do this. First, Zhang introduces
a distance function similar to Equation~\ref{simfunc}, in which
$W(f{_i}) = 1.0$ for all features (i.e., flat Euclidean distance
rather than information-gain weighted distance), in which the distance
between two instances $X$ and $Y$ is normalised by dividing the summed
squared distance by $n$, the number of features, and in which
$\delta(x_{i},y_{i})$ is given as Equation~\ref{simpdist}. The
normalised distance function used by Zhang is given in
Equation~\ref{zhangdist}.
\begin{equation}
\Delta(X,Y)=\sqrt{\frac{1}{n} \sum_{i=1}^{n} ( W(f{_i}) \delta(x_{i},y_{i}))^{2} }
\label{zhangdist}
\end{equation}

The intra-concept similarity\index{similarity!intra-concept} of
instance $X$ with classification $C$ is its similarity (i.e.,
$1-$distance) with all instances in the data set with the same
classification $C$: this subset is referred to as $X$'s {\em family},
$Fam(X)$. Equation~\ref{intra} gives the intra-concept similarity
function $Intra(X)$ ($|Fam(X)|$ being the number of instances in $X$'s
family, and $Fam(X)^{i}$ the $i$th instance in that family).
\begin{equation}
Intra(X)=\frac{1}{|Fam(X)|} \sum_{i=1}^{|Fam(X)|} 1.0 - \Delta(X,Fam(X)^{i}) 
\label{intra}
\end{equation}
All remaining instances
belong to the subset of unrelated instances, $Unr(X)$. The
inter-concept similarity\index{similarity!intra-concept} of an
instance $X$, $Inter(X)$, is given in 
Equation~\ref{inter} (with $|Unr(X)|$ being the number of instances
unrelated to $X$, and $Unr(X)^{i}$ the $i$th instance in that subset).
\begin{equation}
Inter(X)=\frac{1}{|Unr(X)|} \sum_{i=1}^{|Unr(X)|} 1.0 - \Delta(X,Unr(X)^{i}) 
\label{inter}
\end{equation}
The typicality of an instance $X$, $Typ(X)$, is the quotient of $X$'s
intra-concept similarity and $X$'s inter-concept similarity, as given
in Equation~\ref{typicalityfunc}.
\begin{equation}
Typ(X)=\frac{Intra(X)}{Inter(X)}
\label{typicalityfunc}
\end{equation}

An instance type is typical when its intra-concept similarity is larger
than its inter-concept similarity, which results in a typicality
larger than 1. An instance type is atypical when its intra-concept
similarity is smaller than its inter-concept similarity, which results
in a typicality between 0 and 1. Around typicality value 1, instances
cannot be sensibly called typical or atypical;
\cite{zhang-typical} refers to such instances as {\em boundary}\/
instances.

In our experiments, we compute the typicality of all instance types in
the training set, order them on their typicality, and remove 1\%, 2\%,
5\%, and 10\% of the instance types with the lowest typicality, i.e.,
the most atypical instance types. In addition to these four
experiments, we performed an additional eight experiments using the
same percentages, and editing on the basis of (i) instance types'
typicality (by ordering them in reverse order) and (ii) their
indifference towards typicality or atypicality (i.e., the closeness of
their typicality to 1.0, by ordering them in order of the absolute
value of their typicality subtracted by 1.0). The experiments with
removing typical and boundary instance types provide interesting
comparisons with the more intuitive editing of atypical instance
types.

Table~\ref{typ-examples} provides examples of four atypical,
boundary, and typical instance types found in the training
set. Globally speaking, (i) the set of atypical instances tend to
contain foreign spellings of loan words; (ii) there is no clear
characteristic of boundary instances; and (iii) instance types with
high typicality values often involve instance types of which the
middle letters are at the beginning of words or immediately following
a hyphen, or high-frequency instance types, or instance types mapping
to a low-frequency class that always occurs with a certain spelling
(class frequency is not accounted for in Zhang's metric).

\begin{table*}
\begin{center}
\begin{tabular}{||cc|r||cc|r||cc|r||}
\hline
\multicolumn{9}{||c||}{instance types} \\
\multicolumn{3}{||c||}{atypical} &
\multicolumn{3}{|c||}{boundary} &
\multicolumn{3}{|c||}{typical} \\
feature values & class & typicality &
feature values & class & typicality &
feature values & class & typicality \\
\hline
{\sf ureaucr}    & 0\schwa \niupsilon & 0.428 & 
{\sf cheques}    & 0ks                & 1.000 & 
{\sf \_\_\_oilf} & 1\openo \i         & 7.338 \\
{\sf freudia}    & 0\openo \i         & 0.442 & 
{\sf elgium\_}   & 0-                 & 1.000 &  
{\sf etectio}    & 0k\esh             & 8.452 \\
{\sf \_tissue}   & 0\esh              & 0.458 & 
{\sf laby\_\_\_} & 0a\i               & 1.000 & 
{\sf ow-by-b}    & 0b                 & 9.130 \\
{\sf \_\_czech}  & 0-                 & 0.542 & 
{\sf manna\_\_}  & 0-                 & 1.000 &
{\sf ng-iron}    & 2a\i \schwa        & 12.882 \\
\hline
\end{tabular}
\caption{Examples of atypical (left), boundary (middle), and
typical (left) instance types in the training set. For each instance
(seven letters and a class mapping to the middle letter), its
typicality value is given.\label{typ-examples}}
\end{center}
\end{table*}

\subsection{Class-prediction strength}

A second estimate of exceptionality is to measure how well an instance
type predicts the class of all instance types within the training set
(including itself). Several functions for computing class-prediction
strength have been proposed, e.g., as a criterion for removing
instances in memory-based ($k$-nn) learning algorithms, such as {\sc
ib3} \cite{aha-ibl} (cf. earlier work on edited $k$-nn
\cite{wilson-edited,voisin-edited}); or for weighting instances in the
{\sc Each} algorithm \cite{salzberg-each,cost-pebls}. We chose to
implement the straightforward class-prediction strength function as
proposed in \cite{salzberg-each} in two steps. First, we count (a) the
number of times that the instance type is the nearest neighbour
of another instance type, and (b) the number of occurrences that when
the instance type is a nearest neighbour of another instance type, the
classes of the two instances match. Second, the instance's
class-prediction strength is computed by taking the ratio of (b) over
(a). An instance type with class-prediction strength 1.0 is a perfect
predictor of its own class; a class-prediction strength of 0.0
indicates that the instance type is a bad predictor of classes of
other instances, presumably indicating that the instance type is
exceptional.

We computed the class-prediction strength of all instance types in the
training set, ordered the instance types according to their strengths,
and created edited training sets with 1\%, 2\%, 5\%, and 10\% of the
instance types with the lowest class prediction strength removed,
respectively. In Table~\ref{pred-examples}, four sample instance types
are displayed which have class-prediction strength 0.0, i.e., the
lowest possible strength. They are never a correct nearest-neighbour
match, since they all have counterpart types with the same feature
values. For example, the letter sequence {\sf \_\_\_algo} occurs in
two types, one associated with the pronunciation /'\ae / (viz.,
primary-stressed /\ae /, or 1\ae \ in our labelling), as in {\sf
algorithm} and {\sf algorithms}; the other associated with the
pronunciation /''\ae / (viz. secondary-stressed /\ae / or 2\ae), as in
{\sf algorithmic}. The latter instance type occurs less frequently
than the former, which is the reason that the class of the former is
preferred over the minority class. Thus, an ambiguous type with a
minority class (a {\em minority ambiguity}) can never be a correct
predictor, not even for itself, when using {\sc ib1-ig} as a
classifier, which always prefers high frequency over low frequency in
case of ties.

\begin{table}
\begin{center}
\begin{tabular}{||cc|r||}
\hline
feature values & class & cps \\
\hline
{\sf \_\_\_algo} & 2\ae               & 0.0 \\
{\sf ck-benc}    & 1b                 & 0.0 \\
{\sf erby\_\_\_} & 0a\i               & 0.0 \\
{\sf reface\_}   & 0e\i               & 0.0 \\ 
\hline
\end{tabular}
\caption{Examples of instance types with the lowest possible class
prediction strength (cps) 0.0.\label{pred-examples}}
\end{center}
\end{table}

\subsection{Friendly-neighbourhood size}

A third estimate for the exceptionality of instance types is counting
by how many nearest neighbours of the same class an instance type is
surrounded in instance space. Given a training set of instance types,
for each instance type a ranking can be made of all of its nearest
neighbours, ordered by their distance to the instance type. The number
of nearest-neighbour instance types in this ranking with the same
class, henceforth referred to as the friendly-neighbourhood size, may
range between 0 and the total number of instance types of the same
class. When the friendly neighbourhood is empty, the instance type
only has nearest neighbours of different classes. The
argumentation to regard a small friendly neighbourhood as an
indication of an instance type's exceptionality, follows from the same
argumentation as used with class-prediction strength: when an instance
type has nearest neighbours of different classes, it is vice versa a bad
predictor for those classes. Thus, the smaller an instance type's
friendly neighbourhood, the more it could be regarded exceptional.

Friendly-neighbourhood size and class-prediction strength are related
functions, but differ in their treatment of class ambiguity. As stated
above, instance types may receive a class-prediction strength of 0.0
when they are minority ambiguities. Counting a friendly neighbourhood
does not take class ambiguity into account; each of a set of ambiguous
types necessarily has no friendly neighbours, since they are
eachother's nearest neighbours with different classes. Thus,
friendly-neighbourhood size does not discriminate between minority and
majority ambiguities. In Table~\ref{fam-examples}, four sample
instance types are listed with friendly-neighbourhood size 0. While
some of these instance types without friendly neighbours in the
training set (perhaps with friendly neighbours in the test set) are minority
ambiguities (e.g., {\sf \_\_\_edib} 2\niepsilon), others are majority
ambiguities (e.g., {\sf \_\_\_edib} 1\niepsilon), while others are not
ambiguous at all but simply have a nearest neighbour at some distance
with a different class (e.g., {\sf soir\'{e}e\_} 0r).

\begin{table}
\begin{center}
\begin{tabular}{||cc|r||}
\hline
feature values & class & fns \\
\hline
{\sf \_\_\_edib} & 2\niepsilon        & 0 \\
{\sf \_\_\_edib} & 1\niepsilon        & 0 \\
{\sf echnocr}    & 1n                 & 0 \\
{\sf soir\'{e}e\_} & 0r               & 0 \\ 
\hline
\end{tabular}
\caption{Examples of instance types with the lowest possible
friendly-neighbourhood size (fns) 0, i.e., no friendly
neighbours.\label{fam-examples}} 
\end{center}
\end{table}

\section{Results}
\label{results}

\begin{figure*}
\centerline{
        \epsfxsize=\textwidth
        \epsfbox{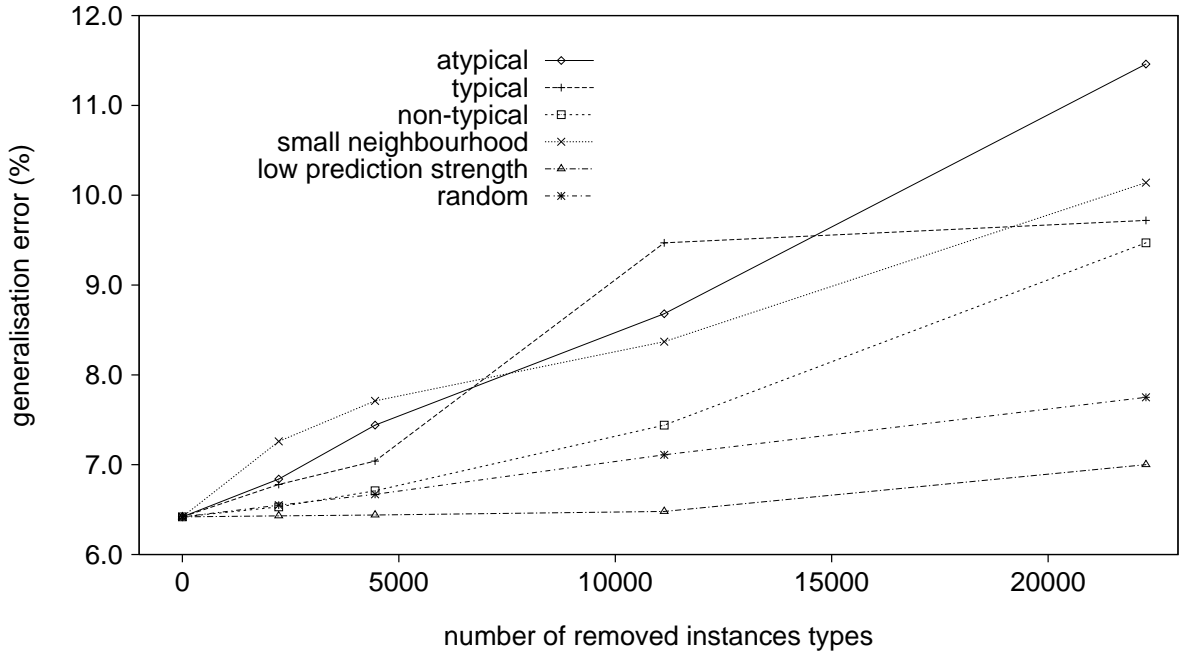}
}
\center
\caption{Generalisation errors (percentages of incorrectly classified
test instances of {\sc tribl-ig}, with increased numbers of edited
instances, according to the tested exceptionality criteria atypical,
typical, boundary, small neighbourhood, low prediction strength,
and random selection. Performances, denoted by points, are measured when
1\%, 2\%, 5\%, and 10\% of the most exceptional instance types are edited.}
\label{edited-gs}
\end{figure*}

Figure~\ref{edited-gs} displays the generalisation accuracies in terms
of incorrectly classified test instances obtained with all performed
experiments. The leftmost point in the Figure, from which all lines
originate, indicates the performance of {\sc ib1-ig} when trained on
the full data set of 222,601 types, viz. 6.42\% incorrectly classified
test instances (when computed in terms of incorrectly pronounced test
words, {\sc ib1-ig} pronounces 64.61 of all test words flawlessly).

The line graph representing the four experiments in which instance
types are removed randomly can be seen as the baseline graph. It can be
expected that removing instances randomly leads to a degradation of
generalisation performance. The upward curve of the line graph
denoting the experiments with random selection indeed shows degrading
performance with increasing numbers of left-out instance types. The
relative decrease in generalisation accuracy is 2.0\% when 1\% of the
training material is removed randomly, 3.8\% with 2\% random removal, 10.7\%
with 5\% random removal, and 20.7\% with 10\% random removal.

Surprisingly, the only experiments showing lower performance
degradation than removal by random selection are those with
class-prediction strength; the other criteria for removing exceptional
instances lead to worse degradations. It does not matter whether
instance types are removed on grounds of their typicality: apparently,
a markedly low, neutral, or high typicality value indicates that the
instance type is (on average) important, rather than removable. The
same applies to friendly-neighbourhood size: instances with small
neighbourhood sizes appear to contribute significantly to performance
on test material. It is remarkable that the largest errors with 1\%
and 2\% removal are obtained with the friendly-neighbourhood size
criterion: it appears that on average, the instances with few or no
nearest neighbours are important in the classification of test
material.

When using class-prediction strength as removal criterion, performance
does not degrade until about 5\% of the instance types with the lowest
strength are removed from memory. The reason is that class-prediction
strength is the only criterion that detects minority ambiguities,
i.e., instance types with prediction strength 0.0, that cannot
contribute to classification since they are always overshadowed by
their counterpart instance types with majority classes, even for their
own classification. In the training set, 9,443 instance types are
minority ambiguities, i.e., 4.2\% of the instance types (accounting
for 3.8\% of the instance tokens in the original token set).

Thus, among the tested methods for reducing the memory needed for
storing an instance base in memory-based learning, only two are
performance-preserving while accounting for a substantial reduction in
the amount of memory needed by {\sc ib1-ig}:

\begin{enumerate}
\item
Replacing instance tokens by instance types accounts for a reduction
of about 63\% of memory needed to  store instances, excluding the memory
needed to store frequency information. When frequency information is
stored in two bytes per instance type, the memory reduction is
about 54\%.
\item
Removing instance types that are minority ambiguities on top of the
type/token-reduction accounts only for an additional memory reduction
of 2\%, i.e., for a total memory reduction of 65\%;
56\% with two-byte frequency information stored per instance. 
\end{enumerate}

\section{Discussion and future research}
\label{discussion}

As previous research has suggested
\cite{daelemans-harmful,daelemans-overview,vandenbosch-thesis},
keeping full memory in memory-based learning of word pronunciation
strongly appears to yield optimal generalisation accuracy. The
experiments in this paper show that optimisation of memory use in
memory-based learning while preserving generalisation accuracy can
only be performed by (i) replacing instance tokens by instance types
with frequency information, and (ii) removing minority
ambiguities. Both optimisations can be performed straightforwardly;
minority ambiguities can be traced with less effort than by using
class-prediction strength. Our implementation of {\sc ib1-ig}
described in \cite{daelemans-hyphen,daelemans-air} already makes use
of this knowledge, albeit partially (it stores class distributions
with letter-window types). We note that with $k>2$ (in our tested
implementation of {\sc ib1-ig}, $k=1$), removing minority ambiguities
may distort performance, since taking more than single nearest
neighbours into account may allow for minority ambiguities to play a
constructive role in classification.

Our results also show that a\-typic\-al\-ity, non-typic\-al\-ity, and
typic\-al\-ity
\cite{zhang-typical}, and friendly-neighbourhood size are all estimates
of exceptionality that indicate the importance of instance types for
classification, rather than their removability. As far as these
estimates of exceptionality are viable, our results suggest that
exceptions should be kept in memory and not be thrown away.

The results of the present study suggest that the following questions
be investigated in future research:

\begin{itemize}
\item
The tested criteria can be employed as instance weights as in {\sc
Each} \cite{salzberg-each} and {\sc Pebls} \cite{cost-pebls}, rather
than as criteria for instance removal. Instance weighting may add
relevant information to similarity matching, and may improve {\sc
ib1-ig}'s performance rather than just preserving it.
\item
The tested implementation of {\sc ib1-ig} performs a $k$-nn search
through instance space with $k=1$. When $k>1$, {\sc ib1-ig}'s
performance may change, as well as the effect of applying the
instance-removal techniques tested here. Removing instances from
memory may have a less drastic effect when more instance types at more
distances are allowed to match a new instance.
\end{itemize}

\subsection*{Acknowledgements}
We thank the members of the ILK group, Ton Weij\-ters, and Eric Postma
for fruitful discussions.

\end{document}